\documentclass{svproc}
\usepackage{url}
\usepackage{graphicx}
\usepackage{verbatim}
\usepackage{xcolor}
\usepackage{float}

\newcommand{\ar}[1]{{\textcolor{black}{{#1}}}}

\newcommand{\virgolette}[1]{``#1''}
\usepackage{colortbl}
\usepackage{cite}

\begin{document}
\mainmatter              
\title{Systematic analysis of requirements \\for socially acceptable service robots}
\titlerunning{Systematic analysis of requirements}  
%


\author{Andrea Ruo\inst{1} \and Simone Arreghini\inst{2} \and Luca Capra\inst{3} \and Rosario De Chiara\inst{4} \and Valeria Di Pasquale\inst{4} \and Alessandro Giusti\inst{2} \and Cristina Iani\inst{1} \and Antonio Paolillo\inst{2} \and Dominic Petrak\inst{5} \and Alexander Plaum\inst{6} \and Megha Quamara\inst{7} \and Lorenzo Sabattini\inst{1} \and  Viktor Schmuck\inst{7} \and Paolo Servillo\inst{4} \and Francesco Zurolo\inst{4} \and Valeria Villani\inst{1}}

\authorrunning{Andrea Ruo\inst{1} et al.} 
%
\tocauthor{Andrea Ruo, Lorenzo Sabattini, and Valeria Villani}

\institute{{University of Modena and Reggio Emilia, Reggio Emilia, Italy}\\
\email{\{name.surname\}@unimore.it}
\and
Dalle Molle Institute for Artificial Intelligence (IDSIA), USI-SUPSI, Lugano, Switzerland,\\
\email{\{name.surname\}@supsi.ch}
\and
Spindox Labs srl, Trento, Italy,\\
\email{luca.capra@spindox.it}
\and
Poste Italiane, Rome, Italy,\\
\email{\{name.surname\}@posteitaliane.it}
\and
Technical University of Darmstadt, Darmstadt, Germany\\
\email{petrak@ukp.informatik.tu-darmstadt.de}
\and
Deutsche Welle, Bonn, Germany,\\
\email{alexander.plaum@dw.com}
\and
King's College London, London, UK,\\
\email{\{name.surname\}@kcl.ac.uk}}

\maketitle              
\begin{abstract}
In modern society, service robots are increasingly recognized for their wide range of practical applications. In large and crowded social spaces, such as museums and hospitals, these robots are required to safely move in the environment while exhibiting user-friendly behavior. Ensuring the safe and socially acceptable operation of robots in such settings presents several challenges. 
\ar{To enhance the social acceptance in the design process of service robots,} we present a systematic analysis of requirements, categorized into functional and non-functional
. These requirements are further classified into different categories, with a single requirement potentially belonging to multiple categories. 
Finally, considering the specific case of a receptionist robotic agent, we discuss the requirements it should possess to ensure social acceptance.
\end{abstract}
\section{Introduction}
\label{sec:introduction}
Service robots offer many potential benefits, such as increased productivity, consistent service quality, and reduced personnel costs. Over the past decade, there has been a substantial increase in applications where robots operate in environments shared with people, known as social spaces \cite{onlineRobotNavigation}.  
Examples of such applications include healthcare \cite{holland2021service}, where receptionist agents assist patients with booking appointments and guiding them to specific offices; museums \cite{draghici2022development}, where they enhance the visitor experience by offering guided tours and interactive exhibit information; or airports \cite{Spencer}, where they can speed up the check-in process, provide real-time flight updates, and assist passengers with directions and general inquiries, all of which play a crucial role in enhancing user experience and operational efficiency.

For instance, Stricker et al. \cite{university} introduced TOOMAS, an interactive reception robot designed for shopping environments. This robot is capable of autonomously approaching potential customers, navigating the marketplace, and guiding customers to their selected products, thereby offering a fully accompanied shopping experience. In \cite{foster2019mummer}, the authors, as part of the EU-funded MuMMER project, developed a social robot designed to interact naturally and flexibly with users in public spaces such as shopping malls. The EU-funded project SPENCER \cite{Spencer} developed a reception robot designed to assist, inform, and guide passengers in large and crowded airports. This robot integrates map representation, laser-based people and group tracking, activity and motion planning to efficiently manage passenger assistance.
Moreover in \cite{ruo2024Follow}, the authors present an architecture for user identification and social navigation using a mobile robot. This architecture leverages computer vision, machine learning, and artificial intelligence algorithms to identify and guide users within a social navigation context, thereby providing an intuitive and user-friendly experience with the robot.


To enhance the diffusion of such systems, it is fundamental to guarantee safety and smooth interaction with the user. As regard safety, the robot must not cause any harm to the user. This topic has been largely investigated in the literature. For example, in the case of collaborative robots, the safety standards and collaborative modes have been considered in \cite{villani2018survey}. In navigation tasks, collisions with the users must be avoided \cite{ruo2024CBF,savkin2015safe}. 
However, to guarantee smooth interaction, treating humans simply as obstacles may not be sufficient. In \cite{fong2003survey}, the authors have considered issues such as human-oriented perception, natural human-robot interaction, readable social cues, and real-time performance, presenting a taxonomy of design methods and system components used to build socially interactive robots.
In \cite{arreghini2024wild} the authors have provided an analysis of users intentions, robot behaviors, and their impact on the interaction, building upon the works in \cite{abbate2024self} and \cite{arreghini2024long}.

Additionally, it is important to guarantee the acceptance of robots. In this regard, it is important that the user feels comfortable interacting with the robots and perceives the system as understandable, pleasant, easy to use, and useful~\cite{davis1989perceived}.
\ar{Wirtz et al. \cite{wirtz2018brave} have presented the Service Robot Acceptance Model (sRAM) to explain how and why consumers accept and use service robots. The authors have argued that consumer acceptance of service robots depends on how well these robots can meet user's needs and achieve role congruence. The sRAM identifies three categories of elements that influence robot acceptance: functional elements (such as perceived ease of use and usefulness), social-emotional elements (such as perceived humanness and social interactivity), and relational elements (such as trust and rapport).
In \cite{chiang2020impacts, liu2023service}, studies have been conducted to evaluate the quality of service based on real data. Specifically, \cite{chiang2020impacts} has found that customers' highest priorities for robot service quality are assurance and reliability, while tangibility and empathy are not as important. Meanwhile, \cite{liu2023service} has found that customers in hospitality and restaurant scenarios have exhibited more unethical behavior when excluded by human staff rather than robots under service exclusion, but more unethical behavior when served by robots compared to human staff in inclusive scenarios.}

This paper aims to provide a thorough discussion of the requirements 
\ar{in order to increase the social acceptance in the design process of service robots}. Perspectively, leveraging such requirements in the design allows human-robot interaction to 
\begin{itemize}
    \item satisfy the goals and needs of the user;
    \item comply with the social context in which the system is being used;
    \item be transparent, safe, secure, explainable, and trusted by the user.
\end{itemize}

The rest of the paper is structured as follows: in Sec.~\ref{sec:social_acceptance}, the concept of social acceptance for service robots is discussed.
Sec.~\ref{sec:requirements} introduces a detailed description of requirements in order to consider a service robot as socially acceptable; Sec.~\ref{sec:implementation_example} presents an implementation example wherein these requirements are applied, and Sec.~\ref{sec:conclusion} summarizes the conclusions of this paper.

\section{Social acceptance in robotics}
\label{sec:social_acceptance}
In the literature, acceptance of a technical system takes on various connotations~\cite{gaudiello2016trust, heerink2010assessing}. It is important to underline that in this context, acceptance is defined at an individual level, i.e., regarding the single user, and not at a societal level. \ar{A systematic review of the literature is presented in \cite{savela2018social}, discussing how social acceptance of robots has been studied in different occupational settings and what kinds of attitudes the studies have uncovered about robots as workers.} In~\cite{beer2011understanding}, the authors define acceptance as the positive evaluation that results in the motivation and eventual act of using the technology for the task it is designed to support. In particular, consider the case of a receptionist agent whose task is to provide information to users and guide them within a social environment. In this scenario, the agent should exhibit specific behaviors that are user-friendly and socially acceptable, to ensure an optimal user experience. 

In the case of social robots, social acceptance may also refer to the capability of the technology to be used in different social contexts in such a way that it does not make users feel uncomfortable or out of place. Indeed, social acceptance may be defined as a user feeling comfortable with an artificial agent as a conversational partner, finding its social skills credible, and accepting social interaction as a way to communicate with it~\cite{heerink2010assessing}. In any use case, there are different aspects that must be considered to ensure the acceptability of a service robot. 

The social acceptance of robotic agents in social spaces, following the connotations defined in~\cite{gaudiello2016trust} and summarized in Table~\ref{tab:acceptance_connotations}, can be divided into different aspects.  
Firstly, physical acceptance refers to how users perceive the agent based on its appearance and physical aspect. For instance, a robot designed with a friendly appearance is more likely to be accepted by users.
Secondly, behavioral acceptance involves user's perceptions of the agent's use of the space and its non-verbal communication. Effective non-verbal cues, such as gestures and facial expressions, can make the interaction appear more believable, fluent, and natural, thus enhancing user experience.
Functional acceptance relates to the agent's practicality, innovativeness, and ease of use. 
Users are more likely to accept a robot that is perceived as useful, accurate, and innovative in its functionality. This includes the robot's ability to perform tasks effectively and efficiently.
In terms of social acceptance, users must perceive the robot as a social entity capable of engaging in social behaviors. The satisfaction of users is also influenced by the opinions of others regarding human-machine interactions.
The union between functional and social acceptance gives rise to trust, that is, the belief that the agent acts with integrity and reliability.
Cultural acceptance addresses how well the agent aligns with the user's cultural norms and values. This includes factors like educational values and the general tech-savviness of the target audience. An agent that respects and reflects cultural aspects is more likely to be accepted by users.
Lastly, representational acceptance is about users viewing the agent positively overall. This involves a general perception of the robot's role and presence in their environment.



\begin{table}
    \caption{Acceptance connotations \cite{gaudiello2016trust}}
    \begin{center}
        \begin{tabular}{l@{\quad}p{7cm}}
        \hline
        \multicolumn{1}{l}{\rule{0pt}{12pt}Requirement}&
        \multicolumn{1}{l}{Description/Rationale}\\[2pt]
        \hline
        \rule{0pt}{12pt}
        \hspace{-0.1cm}Physical acceptance & Users perceive the agent as likable and credible based on its physical aspect\\
        Behavioral acceptance & Users perceive the agent's non-verbal communication as believable and the interaction as fluent, natural, and pleasant\\
        Functional acceptance & Users perceive the agent as easy-to-use, useful, accurate, and innovative\\
        Social acceptance & Users perceive a social entity in the agent, consider it capable of performing social behavior, and are satisfied with what other people think of human-machine interaction\\
        Cultural acceptance & Users accept the agent because it complies with their culture in general (e.g.: its educational values and tech-savviness)\\
        Representational acceptance & Users consider the agent in a positive way \\[2pt]
        \hline
        \end{tabular}
    \end{center}
    \label{tab:acceptance_connotations}
\end{table}

\section{Requirements for socially acceptable service robots}
\label{sec:requirements}
The requirements defined and analyzed in this paper are expected to elicit acceptability (i.e., a priori positive evaluation by users when confronted with the system), which will then lead to general acceptance (i.e., a positive, long-term retrospective evaluation)~\cite{beer2011understanding} of service robots. In more detail, it is possible to divide the requirements into functional and non-functional requirements:
\begin{itemize}
    \item Functional requirements (FRs) capture the intended behavior of the system. This behavior may be expressed as a service or function which the system is required to perform.
    \item Non-Functional requirements (NFRs) specify the criteria used to judge the operation of a system, rather than its specific behavior. 
\end{itemize}

The requirements presented in this section cover all types of acceptance, as schematized in Table~\ref{tab:acceptance_connotations}. In addition, a service robot must also be designed to meet the additional requirements schematized in Table~\ref{tab:acceptance_NFR}. Alongside these requirements, service agents may need additional technical NFRs, which depend on the specific use case.


\begin{table}
    \caption{Additional Requirements}
    \begin{center}
        \begin{tabular}{l@{\quad}p{9cm}}
        \hline
        \multicolumn{1}{l}{\rule{0pt}{12pt}Requirement}&
        \multicolumn{1}{l}{Description/Rationale}\\[2pt]
        \hline
        \rule{0pt}{12pt}
        \hspace{-0.1cm}Compliance & The agent correspond with the rules and the regulation imposed by the use cases\\
        Ethics & The agent is civil, respectful, privacy-friendly, transparent, trustworthy\\
        Performance &  The agent is fast, safe, reliable, robust, scalable\\
        Economics & The agent is innovative, realizable on designated budgets, and potentially marketable\\
        Sustainability & The agent is sufficiently documented, generally compatible and modular\\
        Eco-Friendliness & The agent is non-toxic, as energy-efficient and resource-saving as possible, durable, recyclable \\[2pt]
        \hline
        \end{tabular}
    \end{center}
    \label{tab:acceptance_NFR}
\end{table}
The requirement analysis proposed in this work follows an approach guided by real-world use cases, bottom-up principles, and user-centered design derived from the European project SERMAS. This project aims to develop innovative, formal, and systematic methodologies and technologies for modeling, developing, analyzing, testing, and studying the use of socially acceptable advanced technology system.
From the use cases of SERMAS, user's needs have been identified, leading to the derivation of requirements. 
To clarify the analysis, we can group all requirements into one or more of the following categories: Interaction, Operability, Perception, Environment, Privacy, Regulations and Safety.

\subsection{Interaction}
\label{subsec:interaction}
This category includes requirements related to how users interact with the service robot, encompassing input methods like text input, touchscreen interaction, voice commands, and gestures. Effective interaction mechanisms are crucial for ensuring user-friendly experiences, enabling users to communicate their needs and commands efficiently. The system should support multiple interaction modes to accommodate diverse user preferences and ensure accessibility for individuals with varying abilities. 

\subsection{Operability}
\label{subsec:operability}
Operational requirements in this category pertain to the functionality and stability of hardware and software components, ensuring smooth operation and reliability of the service robot. This encompasses the system's ability to perform tasks consistently over time without failures, as well as its capacity for regular maintenance and updates. High operability is essential for maintaining user trust and ensuring continuous, uninterrupted service.

\subsection{Perception}
\label{subsec:perception}
This category includes requirements concerning the sensory inputs used by the service robot, such as computer vision, sensors, and other data sources for interpreting the environment. Accurate perception is vital for the system to understand and appropriately respond to its surroundings. This involves recognizing objects, interpreting human actions and emotions, and making informed decisions based on real-time data from various sources.

\subsection{Environment}
\label{subsec:environment}
This category encompasses requirements addressing the environment in which the service robot operates, including both physical and digital environments. The agent must demonstrate adaptability to different environmental conditions and seamless integration with existing infrastructures. These environmental conditions may include factors such as lighting, noise, object arrangement, and so on.

\subsection{Privacy}
\label{subsec:privacy}
Privacy requirements involve measures to protect sensitive data, including personal and confidential information, ensuring proper handling and compliance with privacy regulations. The agent must implement robust data security protocols to prevent unauthorized access and misuse of information. Ensuring privacy protection is fundamental to building user trust and complying with legal standards.

\subsection{Regulations}
\label{subsec:regulations}
Regulatory requirements encompass adherence to company policies, industry standards, and legal regulations governing the use and deployment of service robots. Compliance with these regulations is crucial to avoid legal repercussions and ensure ethical use of technology.

\subsection{Safety}
\label{subsec:safety}
Safety requirements ensure that the agent does not pose risks or hazards to users, minimizing the potential for injury or trauma during operation. The system should be designed with safety features to prevent accidents and respond effectively to emergencies. 

\begin{figure}
    \centering
    \includegraphics[width=\textwidth]{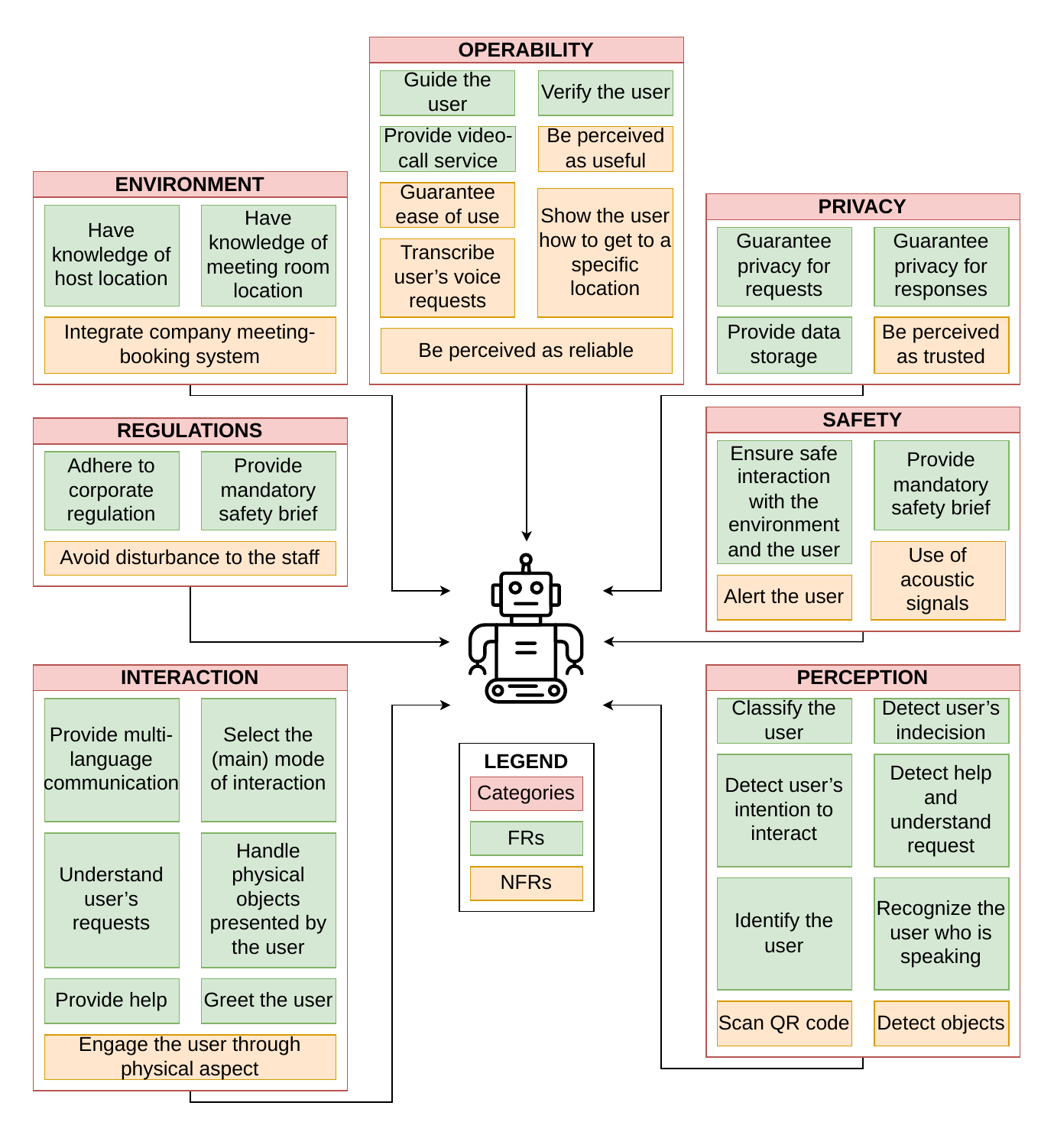}
    \caption{Requirements necessary for a socially acceptable receptionist agent are categorized into FRs, depicted in green, and NFRs, depicted in orange, which can be grouped into various categories, shown in red.}
    \label{fig:RecAgentReq}
\end{figure}

\section{Receptionist robot: a case study}
\label{sec:implementation_example}
The proposed requirement analysis allows the development of service robots that are socially acceptable, through the selection of specific FRs or NFRs, as explained in Sec.~\ref{sec:requirements}.

As discussed in Sec.~\ref{sec:introduction}, there can be various application contexts for a service agent, such as a bank, airport, museum, hotel, or convention center.
In this work, we focus on the case of a mobile service robot, which serves as a receptionist agent within a social context. For the system to be considered socially acceptable, it should meet FRs and NFRs, as shown in Fig.~\ref{fig:RecAgentReq}, with the NFRs depending on the particular use case in which the system is deployed.

Within a social environment, a receptionist agent should be able to receive and interact with users based on conversational activities, provide them with information, and guide them to certain destinations safely and acceptably in all its connotations. Furthermore, the agent should include an identification and authorization process to admit users to the building.

In more detail, the service robot must be capable of providing users with various types of interaction services, as each user may have different characteristics. In fact, users may be adults or children, have different nationalities, level of experience with digital services, or disabilities. To achieve this, we can define the following FRs and NFRs:
\begin{itemize}
    \item \textbf{Interaction FRs:}
    \begin{itemize}
        \item \textbf{Provide multi-language communication}: users will have the possibility to choose their preferred language for communications, enabling interaction adaptation based on the user's language preference.
        \item \textbf{Greet the user}: the service robot should greet the user using the most appropriate interaction style for the identified user class.
        \item \textbf{Understand user's requests}: the agent should be able to understand various requests, such as accessing the building, locating specific areas within the building, accessing security information, or receiving directions to particular locations within the building, all aimed at meeting user needs.
        \item \textbf{Provide help}: the agent should be designed to offer useful suggestions and guidance for resolving common issues or assisting users when they encounter difficulties. Additionally, it should present a list of services it can provide assistance with, such as accessing a building for a meeting, offering directions to a designated meeting room or office, and providing safety information.
        \item \textbf{Select the (main) mode of interaction}: the system should offer users a choice of text/touch, voice, and/or gesture interaction methods to tailor interaction to individual preferences.
        \item \textbf{Handle physical objects presented by the user}: the agent should have the ability to interact with objects presented by the user to enhance user interaction and engagement. 
    \end{itemize}
    \item \textbf{Interaction NFRs:}
    \begin{itemize}
        \item \textbf{Engage the user through physical aspect}: the service robot should incorporate its own on-screen representation to enhance user engagement. The agent representation will be designed to create a pleasant experience, communicating through voice messages synchronized with facial expressions and body movements. The objective is to convey attentiveness and helpfulness in providing service, all aimed at facilitating effective communication and user engagement.
    \end{itemize}
\end{itemize} 

The receptionist agent must be able to guide the visitor within the social environment until they reach the destination. During this task, the service robot should be able to orient itself, follow the best route to the goal pose, and alert the visitor if he or she has taken the wrong path. To achieve this, we can define the following FRs and NFRs:
\begin{itemize}
    \item \textbf{Operability FRs:}
    \begin{itemize}
        \item \textbf{Provide video-call service}: if required by the application context, the service robot should have the capability to notify the host(s) of a visitor's arrival and initiate a video call (preferably utilizing a tool tailored to the company's requirements) to get approval for access. 
        \item \textbf{Verify the user}: the agent should integrate a user verification procedure (e.g., email link, SMS, etc.).
        \item \textbf{Guide the user}: the service robot should be able to accompany visitors within the environment, retrieve the target destination, orient itself, select the optimal route, and alert visitors if they deviate from the route, prioritizing safety by avoiding obstacles.
    \end{itemize}
    \item \textbf{Operability NFRs:}
    \begin{itemize}
        \item \textbf{Guarantee ease of use}: the user should perceive the system easy to use and comprehend, aiming to enhance user satisfaction and minimize errors.
        \item \textbf{Be perceived as useful}: the user must perceive the agent as useful, to increase the likelihood of its use.
        \item \textbf{Be perceived as reliable}: the user should have the perception of reliability of the service robot, to increase the probability of its use.
        \item \textbf{Show the user how to get to a specific location}: the agent should be able to display a map and provide guidance to the user on navigating from point A to point B to assist users in navigating the social environment.
        \item \textbf{Transcribe user's voice requests}: the service robot should possess the capability to convert user's spoken commands into written text, utilizing functionalities such as speech-to-text (STT), to facilitate user interaction and data processing. 
    \end{itemize}
\end{itemize}

Furthermore, the receptionist agent should be able to autonomously detect a user approaching with the intention to interact with the system. Additionally, the service robot should be able to classify the user based on significant features extracted from the detected approaching visitor. This is useful for the adaptation of social interaction. To achieve this, we can define the following FRs and NFRs:
\begin{itemize}
    \item \textbf{Perception FRs:}
    \begin{itemize}
        \item \textbf{Detect user's intention to interact}: the agent should autonomously detect approaching visitors, defined as individuals who move toward and stop in front of it, excluding passing visitors, to initiate interaction and ensure secure building access.
        \item \textbf{Classify the user}: the service robot should be able to classify a visitor based on key characteristics extracted from the detected approaching visitor (e.g., age), facilitating the adaptation of social interaction to the specific user.
        \item \textbf{Detect user's indecision}: the agent should be able to perceive the user's indecision during use and should possess a mechanism to suggest an assistance function to the user so as to help him or her.
        \item \textbf{Detect help and understand request}: the service robot should recognize when a visitor explicitly asks for assistance in understanding his or her request, whether it is expressed in  language, or through gestures, to improve user understanding.
        \item \textbf{Identify the user}: the system should be capable of recognizing the presence of a user to initiate interaction, following the procedure outlined in corporate regulations.
        \item \textbf{Recognize the user who is speaking}: the service robot will have the capability to identify the user who is speaking, particularly in scenarios where multiple users are present, to initiate interaction with the appropriate user.
       
    \end{itemize}
    \item \textbf{Perception NFRs:}
    \begin{itemize}
        \item \textbf{Scan QR code}: if required by the application context, the system should be capable of scanning the QR code provided by the user and verifying the contained information.
        \item \textbf{Detect objects}: the service agent should identify whether the user is holding an object (such as a parcel), and the interaction is initiated only at the user's request, to offer support when requested.
    \end{itemize}
\end{itemize}

In addition, during the tasks performed by the service robot within the case study, it is necessary for the receptionist agent to be able to adapt to different environmental conditions with existing infrastructures and be familiar with the surrounding environment. For this reason, we can
define the following FRs and NFRs:
\begin{itemize}
    \item \textbf{Environment FRs:}
    \begin{itemize}
        \item \textbf{Have knowledge of host location}: the service robot should have a mapping of the building employees and their location.
        \item \textbf{Have knowledge of meeting room location}: if required by the application context, the agent should have a mapping of the meeting rooms and their location.
    \end{itemize}
    \item \textbf{Environment NFRs:}
    \begin{itemize}
        \item \textbf{Integrate company meeting-booking system}: if required by the application context, the system should be capable of verifying meeting approval.
    \end{itemize}
\end{itemize}

Potential weaknesses in the design or implementation of the service robot could be exploited by malicious actors or regular human users during human-agent interactions, leading to security breaches or agent compromise. Additionally, users may raise objections during the case study execution phase, fearing issues related to the insecure management of collected data. For this reason, we can define the following FRs and NFRs:
\begin{itemize}
    \item \textbf{Privacy FRs:}
    \begin{itemize}
        \item \textbf{Guarantee privacy for requests}: the agent should ensure privacy for requests, aligning with both legal mandates and company policies.
        \item \textbf{Guarantee privacy for responses}: the service robot should ensure privacy for responses, aligning with both legal mandates and company policies.
        \item \textbf{Provide data storage}: data used for body tracking, facial landmarks, or any other information classified as \virgolette{personal data} must be anonymized and not retained, in line with legal and corporate regulations. 
    \end{itemize}
    \item \textbf{Privacy NFRs:}
    \begin{itemize}
        \item \textbf{Be perceived as trusted}: the user must be assured that interactions with the agent are safe with regard to the processing of personal data or, at the very least, the perceived risk associated with inappropriate data handling must be minimal in order to increase the likelihood of its use. 
    \end{itemize}
\end{itemize}

In addition, the service robot must adhere to corporate security and safety regulations to meet legal and company requirements. To
achieve this, we can define the following FRs and NFRs:

\begin{itemize}
    \item \textbf{Regulations FRs:}
    \begin{itemize}
        \item \textbf{Adhere to corporate regulation}: the service robot must adhere to corporate security and safety regulations to meet legal and company requirements.
        \item \textbf{Provide mandatory safety brief}: the agent should provide mandatory information and/or company regulation on safety that are necessary for access to the building, before enabling the visitor to access.
    \end{itemize}
    \item \textbf{Regulations NFRs:}
    \begin{itemize}
        \item \textbf{Avoid disturbance to the staff}: the system must not disrupt or intrude upon the tasks performed by reception employees, in accordance with company requirements. 
    \end{itemize}
\end{itemize}

Lastly, the service robot, during the execution of tasks within the case study, must ensure safety. To achieve this, we can define the following FRs and NFRs:

\begin{itemize}
    \item \textbf{Safety FRs:}
    \begin{itemize}
        \item \textbf{Ensure safe interaction with the environment and the user}: the service robot must be designed to interact safely with both its surroundings and the people it serves, minimizing the risk of accidents or injuries.
        \item \textbf{Provide mandatory safety brief}: the agent should provide visitors with essential information on security and/or business regulations required for access to the building before granting them entry, ensuring compliance with organizational guidelines. 
    \end{itemize}
    \item \textbf{Safety NFRs:}
    \begin{itemize}
        \item \textbf{Alert the user}: the system should have the capability to alert human operators to abnormal environmental parameters, aiming to enhance environmental quality.
        \item \textbf{Use of acoustic signals}: the service robot should be equipped with an audible signal or similar measure that activates during movement to signal its presence to people.
    \end{itemize}
\end{itemize}

\section{Conclusions}
\label{sec:conclusion}
Nowadays, service robots are increasingly recognized for their wide range of practical applications in environments shared with people. This necessitates ensuring that service robots are not only satisfactory, transparent, safe, secure, explainable, and trustworthy but also sustainable. The lack of social acceptance is seen as a barrier to the adoption of service robots and may increase social inequality.

In this work, a systematic analysis of the requirements for achieving acceptability in service robots was conducted. To simplify the analysis, we have categorized the requirements into distinct groups, distinguishing between functional and non-functional requirements. Furthermore, we have demonstrated the application of this analysis through the implementation example of a socially acceptable receptionist agent.

\section*{Acknowledgment}
This work was supported by Horizon Europe program under the Grant Agreement 101070351 (SERMAS) and by the Swiss State Secretariat for Education, Research and Innovation (SERI) under contract number 22.00247.
%
%
\bibliographystyle{unsrt}
\bibliography{main}
\end{document}